# Holistic Transformer: A Joint Neural Network for Trajectory Prediction and Decision-Making of Autonomous Vehicles

Hongyu Hu[a], Qi Wang[a], Zhengguang Zhang[a], Zhengyi Li[a] and Zhenhai Gao[a]*

[a] State Key Laboratory of Automotive Simulation and Control, Jilin University, Changchun, 130022 China.
(e-mail: huhongyu@jlu.edu.cn, wqi19@mails.jlu.edu.cn, zgzhang21@mails.jlu.edu.cn, lizy21@mails.jlu.edu.cn, gaozh@jlu.edu.cn)

*Abstract*—Trajectory prediction and behavioral decision-making are two important tasks for autonomous vehicles that require good understanding of the environmental context; behavioral decisions are better made by referring to the outputs of trajectory predictions. However, most current solutions perform these two tasks separately. Therefore, a joint neural network that combines multiple cues is proposed and named as the holistic transformer to predict trajectories and make behavioral decisions simultaneously. To better explore the intrinsic relationships between cues, the network uses existing knowledge and adopts three kinds of attention mechanisms: the sparse multi-head type for reducing noise impact, feature selection sparse type for optimally using partial prior knowledge, and multi-head with sigmoid activation type for optimally using posteriori knowledge. Compared with other trajectory prediction models, the proposed model has better comprehensive performance and good interpretability. Perceptual noise robustness experiments demonstrate that the proposed model has good noise robustness. Thus, simultaneous trajectory prediction and behavioral decision-making combining multiple cues can reduce computational costs and enhance semantic relationships between scenes and agents.

1 INTRODUCTION

The emergence of autonomous vehicles promises to improve people's travel efficiency and safety. Compared with human drivers, autonomous vehicles are supposed to have a more comprehensive and accurate perception of the surrounding environment [1]. Based on social knowledge obtained by perception, autonomous vehicles can better infer the intentions of surrounding agents and decide the subsequent tactical maneuvers. Currently, autonomous vehicles perform reasonable path planning and motion control to realize the tracking of its expected trajectory. In this process, the intention inferencing of surrounding agents and the decision-making of the ego-vehicle are the two key modules that need to function correctly to realize automated driving [2, 3].

In most autonomous driving systems, trajectory prediction and behavior decision-making of the ego-vehicle are applied as two partitioned modules using separate models [4]. However, there is a strong correlation between trajectory prediction and decision-making as both require a full understanding of the surrounding environment, and the decision-making task requires trajectory prediction information. In fact, trajectory prediction by the ego-vehicle outputs a feasible trajectory that contains decision-making information. The combination of decision-making and trajectory prediction will better enable the network to share and learn complementary information. Furthermore, the decision-making task plays the role of regularization, which improves the performance of all tasks [5, 6]. As these tasks are shared, the speed of inferencing is thus improved, alleviating resource limitations currently hampering the performance of autonomous vehicles.

Generally, several features are extracted for trajectory prediction, including dynamic and kinematic features of agents, interaction features among agents, and interaction features between agents and the environment [7]. Decision information is also considered, which requires the neural network to process multiple cues simultaneously. Both correlation and redundancy exist in these cues. The traditional approach for this simultaneous processing is to simply concatenate them for decoding. This approach has two disadvantages. First, when multiple cues are simply concatenated, the decoder often has difficulty making full use of the cues; moreover, it is difficult to eliminate redundant information. Second, performing separate feature extraction procedures is not conducive to understanding the intrinsic relationships among cues [8]. To better concatenate multiple cues, we propose a novel holistic transformer (HT) that selects different attention mechanisms under the conditions of no prior knowledge, partial prior knowledge and posteriori knowledge of decisions. Then, it efficiently selects the useful information from multiple cues and reduces learning difficulty.

The context representations of the surrounding environment are essential to scene understanding; these representations are the premise of trajectory prediction and decision-making. Traditional methods only select a few key features [9], making it difficult to solve the interaction profiles of multiple agents in complex scenarios. Hence, the accuracy is negatively affected by the multimodality of agent behaviors. To represent heterogeneous environment information, some deep learning methods rasterize scene and agent information into image-style inputs to represent lane lines, obstacles, and agents using different colors [10, 11]. These inputs are fed to the backbone for encoding and to obtain a good contextual representation of the environment. However, rasterized images are prone to lossy encoding, and topological relationships between environmental elements are difficult to model explicitly. For example, with the current methods, if a lane of interest extends for a long distance, the rasterized images not only recognize and track it but also track other useless large areas. This not only wastes computational resources but also affects model presentation. To narrow the autonomous vehicle's cues to a region of interest (ROI), this study adopts the vector representation method, which applies vectorized encoding of scenes and agent trajectories and uses their various attributes as vector features [12]. This method can be used to efficiently model the spatial topological relationships between scenes and agents and improve environmental factor encoding efficiency [13].

HT is a novel network that outputs trajectory predictions and behavior decisions simultaneously. The proposed model consists of an encoder, a feature selection network, and a decoder. The encoder is divided into agent encoding and scene encoding modules. Agents lacking prior knowledge are encoded using sparse multi-head self-attention to reduce the impact of perceptual noise. The learning lane graph representations for motion forecasting network (LaneGCN) [13] is used to encode scene information. Subsequently, we provide a new feature selection network based on partial prior knowledge to model the multimodal interactions among agents and between agents and scenes using sparse attention. According to the posteriori knowledge of decision-making, sigmoid-based self-attention is adopted to fuse decision-making information to output multiple trajectories alongside their probabilities. Finally, decision-posterior knowledge attention visualization, perceptual noise robustness, and ablation comparison experiments are performed to demonstrate the advantages of the proposed model.

The main contributions of this study can be summarized as follows:

- A joint neural network (i.e., HT) for simultaneous trajectory prediction and behavioral decision-making based on vector representation, which efficiently models the spatio-temporal topological relationships between scenes and agents and enhances the semantic relationships among vectors
- An attention-weighted regularizer for trajectory prediction that uses the behavior decision task to improve performance and reduce computational costs
- An organically integrated system of attention mechanisms (i.e., sparse multi-head, sparse feature selection, and multi-head with sigmoid) based on the characteristics of multiple cues.

The remainder of this paper is organized as follows. **Section 2** provides a literature review of related works. **Section 2** presents the proposed method, including the network structure and loss function. The experimental settings, results, and evaluations are described in **Section 4**. Finally, **Section 5** presents concluding remarks and limitations and highlights the scope of future works.

## 2 RELATED WORKS

Presently, there are many examples of trajectory prediction and behavioral decision-making studies, and this section examines the ones closely related to the proposed model.

### 2.1 Traditional methods

Traditional prediction methods focus on dynamic and kinematic parameters of the vehicle. As the vehicle yaw rate and acceleration cannot change suddenly in a short period, the constant yaw rate and acceleration (CYRA) model has been adopted most frequently [14]. However, for long-term prediction, CYRA's prediction performance is severely degraded as the assumption of constancy does not hold [7]. To improve the accuracy of long-term prediction, many models have begun considering driving patterns and agent interactions. For example, Houenou et al. [9] proposed a method of combining CYRA with maneuver recognition, which discretizes the solution space into multiple trajectories and selects the optimal one based on road topology. Considering the interaction between agents and the surrounding environment in long-term prediction, Xie et al. [15] proposed a multi-interaction model for short- and long-horizon trajectory prediction. A hidden Markov model was adopted to identify driving maneuvers. Finally, the multiple interacting model combines short-horizon vehicle physical modeling with long-horizon maneuver recognition. Additionally, a Gaussian stochastic process [16] with a hidden Markov model [17] and a finite automatic state machine [18] was applied to trajectory prediction. However, these methods adopt manual features, which poorly model complex scenes and interactions; thus, the multimodality of agent behaviors are not understood accurately.

Traditional autonomous vehicle decision-making methods aim to make safe and reasonable driving decisions based on the surrounding environment and the state of the ego-vehicle. Currently, owing to their good stability and practicability, finite automatic state machines are often used in trajectory decision-making systems [19]. A finite automatic state machine is a mathematical model of discrete input and output systems consisting of a finite number of states. The current state accepts events and generates corresponding actions, causing state transitions. Decision trees are also applied for decision-making methods [20]. These methods suffer from problems like those of trajectory prediction in that complex rules must be formulated in the face of complex scenes, and the ability of scene understanding is poor.

*2.2 Deep learning methods*

For trajectory prediction, there are three mainstream deep learning methods, according to the classification of scene representation. The first method feeds raw input directly into the neural network and is suitable for simple scenarios, such as highways, expressways, etc. Many methods simply encode the distance between the ego-vehicle and the lane lines, but other important information is often omitted. When given simple road structures, these methods do well mainly by focusing on the interactions among agents. For example, Messaoud et al. [2] proposed a long short-term memory (LSTM) encoder/decoder structure based on multi-head attention, which maps the agent position to an attention matrix and visualizes their relative importance around the ego-vehicle. Deo et al. [21] proposed convolution social pooling, which uses social tensors to encode the motions of surrounding agents. Additionally, a graph convolution neural network (CNN) [22] using generative adversarial imitation learning [23] with an LSTM [24] were applied to model various agent interactions. However, due to the lack of comprehensive scene models, these methods have difficulty dealing with complex scenarios.

To represent complex road topologies, some methods were inspired by the ability of CNNs to extract information from images. By establishing a bird's-eye view raster image as input, Visual Geometry Group and residual neural network (ResNet) backbones are used to encode the scenes [25]. With these, the output of the image convolution is used to decode the trajectory of each agent. For example, Zhao et al. [26] proposed multi-agent tensor fusion, encoded the scene information based on a grid graph, and embedded the multi-agent spatio-temporal information into the scene tensor. Tung et al. [27] proposed CoverNet, which defines the trajectory prediction problem as the classification of different trajectory sets, inputting the semantic map and state into the backbone and trajectory cluster generator, respectively. Doing so eliminates impossible trajectories from the cluster and selects the optimal one. Additionally, anchor trajectories based on the prior knowledge of motion constraints are used to alleviate the problem of multimodal predictions [28]. However, the performance of these methods is limited by the spatial resolution of the grid images, which are easily disturbed by irrelevant raster image regions.

Map information contains strong spatial topology data, which are important for prediction in complex scenarios. A vector representation map describes topological information and ignores useless information. Thus, when encountering a roundabout or a long straight road, it can efficiently model topological relationships. For example, VectorNet [12] represents various road components in the form of vectors, which avoids the lossy rendering of grid maps and the convolutional encoding of intensive calculations. Each vector is connected through a graph CNN to model the connection relationship between each vector component. To capture complex road topologies and long-distance dependencies, LaneGCN [13] was proposed in which global information is integrated through the fusion of scenes and agents in the ROI. Additionally, the distributed representations for graph-centric motion forecasting model [29] and the temporal point-cloud network (TPCN) [30] use similar methods for modeling. However, when these methods fuse agent and scene information, they do not carefully select the predicted features from prior knowledge, thus limiting the expressive ability of the model.

Currently, most deep learning decision-making methods do not include prediction tasks. They generally use CNNs to encode and learn surrounding environments using cameras and other sensors to make classification decisions [31]. Additionally, deep reinforcement learning is popular [32]; however, it often relies on experience from its reward function setting, and it has difficulty understanding the complex topology of map information, resulting in decision-jumping in unknown scenarios.

In summary, several models exist for simultaneous prediction and decision-making. However, if decision information and other prior knowledge can be used as additional branches to improve network performance, it would greatly improve the fit between predictions and decisions while reducing calculation costs and jointly improving accuracy.

3 DEVELOPMENT OF PROPOSED MODEL

*3.1 Problem formulation*

In this study, the proposed model input includes scene and agent information. Scene information refers to lane lines, static obstacles, traffic signs, and traffic lights. These items are often combined with high definition (HD) maps, Global Positioning System localization techniques, and other data sources to establish a complete scene. Additionally, agent information includes all agents in the ROI, such as vehicles, non-motor vehicles, and pedestrians.

The scene information set, $\mathcal{S}$, contains road, traffic light, and other ROI data for the ego-vehicle, and the agent information set, $\mathcal{A}$, includes the ego-vehicle, $A_0$, and all perceived agents $\{A_1, A_2, ..., A_{Ns}\}$, where $Ns$ is the number of perceived agents. $A_i \in \mathbb{R}^{f_a \times t_{hst}}$, $i = 0, 1, 2, ..., Ns$, where $f_a$ is the number of agent features, and $t_{hst}$ is the sequence length. The local coordinate is established by taking the position of the ego-vehicle at $t = 0$ as the origin. The positive direction of the $x$-axis is defined as the ego-vehicle driving direction, and the $y$-axis is the direction in which the $x$-axis rotates 90° counterclockwise. The feature of the agent at time $t$ is $a_t = (\Delta x_t^a, \Delta y_t^a, class^a, flag_t^a)^\top$. Note that $t = -t_{hst}+1, -t_{hst}+2, ..., -1, 0$, where $\Delta x_t, \Delta y_t$ is the delta $x$, $y$ coordinate of the agent of time $t$ and $t$-1, $class$ is the type of agent, and $flag_t$ signifies whether the agent is perceived at time $t$. Additionally, the position coordinate of each agent at $t = 0$ is used as input.

The scene information set, $\mathcal{S}$, is divided into a lane vector feature matrix, $M_f$, and a lane vector adjacency matrix set, $\mathcal{S}_a = \{M_p^{1:6}, M_s^{1:6}, M_r, M_l, M_m, M_o\}$. $M_f \in \mathbb{R}^{f_l \times Nl}$, where $f_l$ is the number of lane vector features, and $Nl$ is the number of lane vectors. Lane vector $m_f = (\Delta x^l, \Delta y^l, heading^l, turn^l, traf^l, intersect^l)^\top$, where $\Delta x^l, \Delta y^l, heading^l$, describes the basic information of the lane vector, $turn^l$ designates whether the lane vector turns left or right, $traf^l$ indicates whether the lane vector is restricted due to traffic lights or lane signs, and $intersect^l$ indicates that the lane vector is in a junction. The meaning of each matrix in the lane vector adjacency matrix set, $\mathcal{S}_a$, is explained in **Table 1** and **Figure 1**. *Pairwise* indicates that the relationship appears in pairs with the same attribute, and ***uniqueness*** indicates whether the relationship occurs more than once for the lane vector. Additionally, the position coordinate of lane vectors is used as input.

**Table 1**
Meanings of the different lane vector adjacency matrices

| Item | Meaning | Pairwise | Uniqueness |
|---|---|---|---|
| $M_p^{1:6}$ | Lane vector upstream of the current lane vector according to direction. 1: 6 represent the first to sixth predecessors. | N | N |
| $M_s^{1:6}$ | Lane vector downstream of the current lane vector according to direction. 1: 6 represent the first to sixth successors. | N | N |
| $M_r$ | Along the direction of the lane vector and to its right side. | N | Y |
| $M_l$ | Along the direction of the lane vector and to its left side. | N | Y |
| $M_m$ | If there are multiple predecessors or successors of the current lane vector, then the lane vector and each predecessor or successor *merge* with each other. The agent has multiple inflow sources or multiple outflow choices for the lane vector. | Y | N |
| $M_o$ | If there are other lane vectors except a single direct predecessor or successor within the range in which the lane vector distance is less than $th_o$, the two lane vectors *overlap* with each other. It should be noted that *merge* must be *overlap*, but not vice versa. | Y | N |

The objective here is to predict $K$ trajectories, $\hat{a}_{t,k} = (\Delta \hat{x}_{t,k}, \Delta \hat{y}_{t,k})^\top$, of all agents in the scene and the probability, $p_k$, of each trajectory, where $K$ is the number of motion modalities, $k = 1, 2, ..., K$, $t = 1, 2, ..., t_{fut}$. Additionally, the goal of this study is to output the maneuver decision, $\hat{D}_m$, and the lane decision, $\hat{D}_l$, of all agents. The maneuver decision refers to the tactical intention of the agent for future actions. **Table 2** shows the types of maneuver decisions, where ***priority*** indicates which decision is defined by the current agent on the premise of meeting multiple decision conditions. Among them, "1" has the highest priority, and "6" has the lowest. *Yield*'s decision has higher priority to improve the security of the decision. Lane decision refers to the lane lines to be followed by the agent in the future.

**Table 2**
Meanings of behavioral decisions

| Decision | Symbol | Priority | Meaning |
|---|---|---|---|
| *Yield + Stop* | S | 1 | Stopped owing to the limitations of surrounding agents, traffic lights, etc. |
| *Yield + Nudge / Lane change* | N | 2 | Owing to the limitations of the surrounding agents, obstacles, global routing, and other factors, the heading angle changes greatly. |
| *Yield + Decrease speed* | D | 3 | Owing to the limitations of the surrounding agents, traffic lights, and other factors, the agent needs to slow down but not stop. |
| *Following* | F | 4 | An agent is considered as *Following* when it follows other agents that are ahead along the heading angle. |
| *Ignore* | I | 5 | Under the premise of abiding by traffic rules, the agent can arbitrarily accelerate or decelerate according to the current state, which is considered as *Ignore*. |
| *Unknown* | U | 6 | Owing to the short duration of perception, *Unknown* is defined when the decision is difficult to determine. |

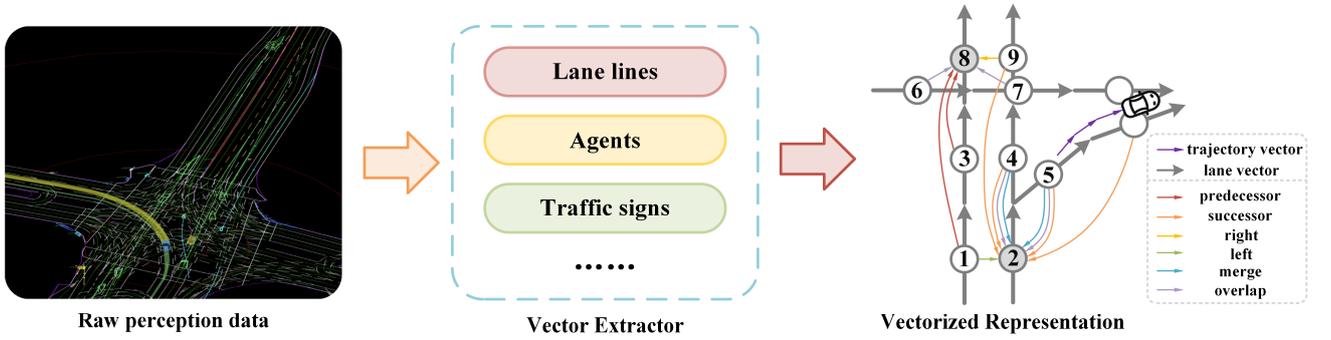

**Figure 1.** Schematic of vectorization feature extraction. For simplicity, only the adjacency relationships of vector 2 and 8 are shown in the figure.

*3.2 Proposed model*

As shown in **Figure 2**, the proposed model includes three parts: an encoder, a feature selection network, and a decoder. These are used to encode agent information and scene information sets, select and fuse multiple cues, and output decoding information. The input of the encoder comes from the input representation module. Each part is described in detail below.

*3.2.1 Encoder*

The encoder encodes the agents and lane vectors and outputs the encoded features for feature selection, cue extraction, and fusion. Because the agent information set is composed of time-series data, and the lane vector set is composed of graphs, there are differences in their encoding methods.

*3.2.1.1 Agent encoding*

Agent trajectory prediction is a type of multivariable time series (MTS) prediction. To encode the MTS, recurrent neural networks (RNNs) are often used to capture temporal features, and LSTMs are more widely used [33]. However, owing to the absence of long-term dependency management, there are weaknesses in the extraction of long-term dependency relationships using RNNs, which can negatively affect trajectory prediction. Additionally, the autonomous vehicle perception module introduces noise to the input data due to severe weather, low-cost sensors, algorithm errors, etc. The traditional approach is to filter out high-frequency noise and smooth the vehicle trajectory. The disadvantage of this is that hand-crafted filtering is added, which may corrupt the original dynamics and kinematics of agents.

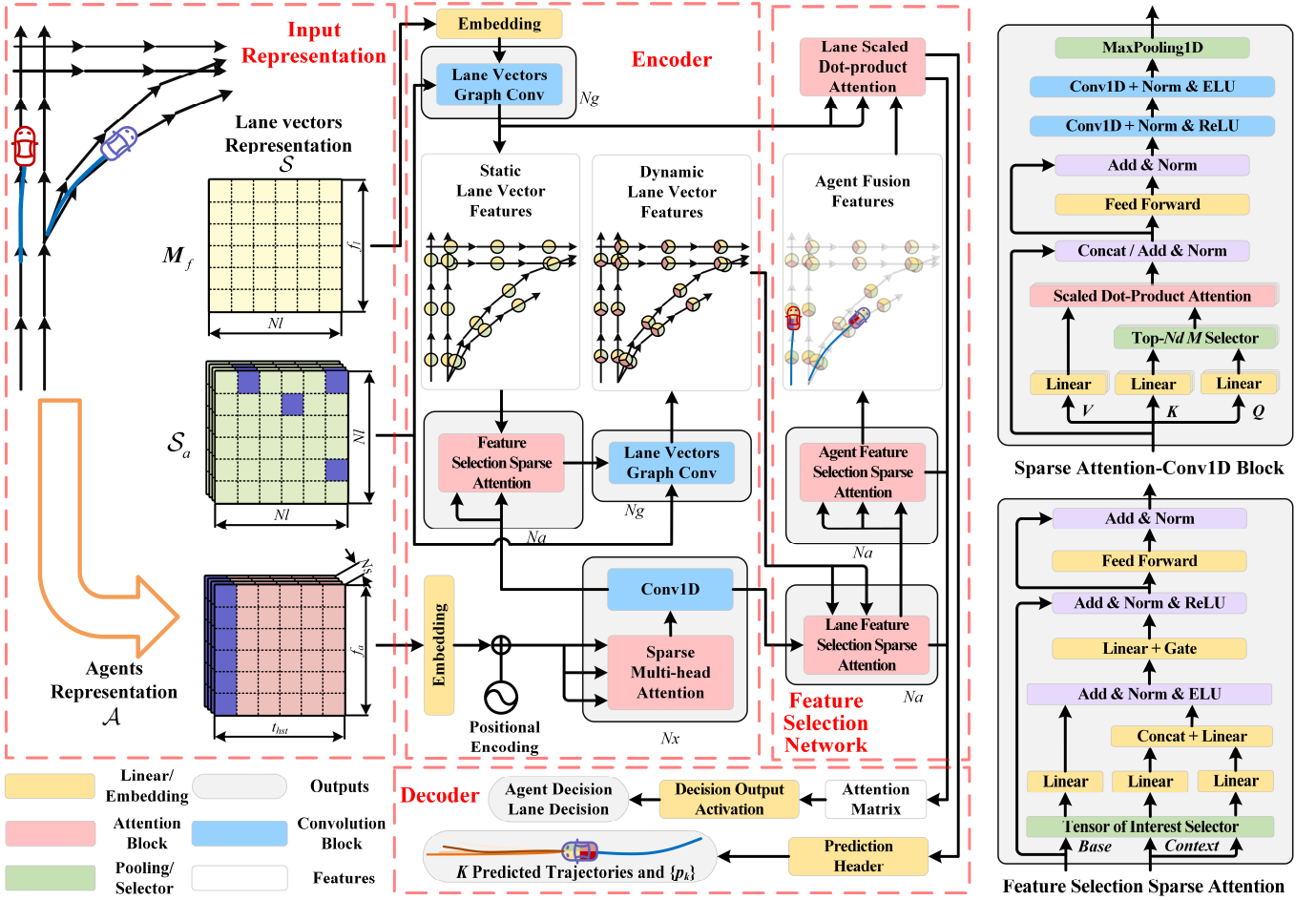

**Figure 2.** Schematic of holistic transformer network framework. The network is mainly composed of the Encoder, Feature Selection Network, and Decoder. The input to the Encoder is shown by Input Representation.

Therefore, two key problems need to be solved in agent encoding: feature extraction across time steps and noise suppression. The vanilla transformer eliminates convolution and extracts features across time steps using multi-head attention. It uses the full connection layer to extract features across multiple time steps. However, it lacks noise suppression. If any time steps contain too much noise, the data will have data-dependent aleatoric uncertainty. To suppress this uncertainty, we must pay less attention to the time steps. Therefore, we use information entropy to measure the uncertainty of each time step to improve its impact with low uncertainty. Inspired by [34] and [35], we modified the vanilla transformer.

Because the transformer adopts scaled dot-product attention and requires global location representation, the positional embedding (PE) layer is used, and the feature dimension after input representation is $d_{model}$,

$$PE_{(pos,2i)} = \sin(pos/(2t_{hst})^{2i/d_{model}}), \tag{1}$$

$$PE_{(pos,2i+1)} = \cos(pos / (2t_{hst})^{2i/d_{model}}),  \tag{2}$$

where $i = 1, 2, \ldots, \lfloor d_{model}/2 \rfloor$.

The standard scaled dot-product self-attention mechanism consists of *query* $\boldsymbol{Q} \in \mathbb{R}^{L \times d}$, *key* $\boldsymbol{K} \in \mathbb{R}^{L \times d}$, and *value* $\boldsymbol{V} \in \mathbb{R}^{L \times d}$, where $L$ is the sequence length, $L=t_{hst}$ in the first sparse attention, and $d$ is the feature dimension of $h$ heads, where $d_{model}=hd$. The input to the attention mechanism is recorded as $X_A$, and $\boldsymbol{Q}$, $\boldsymbol{K}$, and $\boldsymbol{V}$ are obtained through linear layer transformation. The self-attention mechanism is described by the following formula:

$$A(\boldsymbol{Q}, \boldsymbol{K}, \boldsymbol{V}) = \mathrm{Softmax}\left(\frac{\boldsymbol{Q}\boldsymbol{K}^\top}{\sqrt{d}}\right)\boldsymbol{V}. \tag{3}$$

Let $\boldsymbol{q}_i$, $\boldsymbol{k}_i$, and $\boldsymbol{v}_i$ be the vectors of the $i$th row of $\boldsymbol{Q}$, $\boldsymbol{K}$, and $\boldsymbol{V}$, respectively, where $i = 1, 2, \ldots, L$. The attention value for the $i$th in the sequence is

$$A(\boldsymbol{q}_i, \boldsymbol{K}, \boldsymbol{V}) = \sum_j \frac{f(\boldsymbol{q}_i, \boldsymbol{k}_j)}{\sum_k f(\boldsymbol{q}_i, \boldsymbol{k}_k)} \boldsymbol{v}_j, \tag{4}$$

where $f(\boldsymbol{q}_i, \boldsymbol{k}_j) = \exp(\boldsymbol{q}_i \boldsymbol{k}_j^\top / \sqrt{d})$. Note that

$$p(\boldsymbol{k}_j | \boldsymbol{q}_i) = \frac{f(\boldsymbol{q}_i, \boldsymbol{k}_j)}{\sum_k f(\boldsymbol{q}_i, \boldsymbol{k}_k)} \tag{5}$$

is a conditional probability density function, $\sum_j p(\boldsymbol{k}_j | \boldsymbol{q}_i) = 1$. Thus,

$$A(\boldsymbol{q}_i, \boldsymbol{K}, \boldsymbol{V}) = \mathbb{E}_{p(\boldsymbol{k}_j|\boldsymbol{q}_i)}[\boldsymbol{v}_j]. \tag{6}$$

Therefore, the scaled dot-product self-attention is the mathematical expectation of $\boldsymbol{V}$ under distribution $p(\boldsymbol{k}_j | \boldsymbol{q}_i)$. For each sequence step $i$, $\boldsymbol{V}$ is the same, and all uncertainties contained in the current sequence step are reflected in the probability density function $p(\boldsymbol{k}_j | \boldsymbol{q}_i)$. Thus, the smaller the uncertainty of $p(\boldsymbol{k}_j | \boldsymbol{q}_i)$, the better is the outcome. However, we cannot assert what the probability distribution of $p(\boldsymbol{k}_j | \boldsymbol{q}_i)$ should be, nor should we assume what its probability distribution should be, as this is what the neural network must learn. Thus, we look at the problem from the perspective of information entropy. When the information entropy is maximal, its uncertainty is also maximal. Information entropy $\mathcal{H}(q)$ is defined as

$$\mathcal{H}(q) = -\sum q \log q. \tag{7}$$

Finding the maximum value of $H(q)$ is equivalent to

$$\begin{aligned} \min \quad & \sum_{i=1}^L q_i \log q_i \\ \mathrm{s.t.} \quad & \sum_{i=1}^L q_i = 1 \end{aligned} \tag{8}$$

Obviously, because $0 \leq q_i \leq 1$, this problem can be considered a convex optimization problem. Thus, the solution satisfying the Karush–Kuhn–Tucker (KKT) condition is the optimal solution. Under equality constraints, the solution satisfying the KKT condition is found when the first partial derivative of the Lagrange function equals zero. The Lagrange function is given by

$$\mathcal{L}(q_1, q_2, ..., q_L, \lambda) = \sum_{i=1}^{L} q_i \log q_i + \lambda \left( \sum_{i=1}^{L} q_i - 1 \right), \tag{9}$$

where $\lambda$ is the Lagrange multiplier. Then, find the first-order partial derivative of $\mathcal{L}(q_1, q_2, ..., q_n, \lambda)$ with respect to $q_1, q_2, ..., q_n, \lambda$ and set the partial derivative equal to zero:

$$\begin{cases} \lambda = -\log q_1 = -\log q_2 = ... = -\log q_L \\ \sum_{i=1}^{L} q_i = 1 \end{cases}. \tag{10}$$

The problem can then be solved as $q_1 = q_2 = ... = q_L = 1/L$, according to Eq. (10). That is, when $q$ is uniformly distributed as $q \sim U(0, L)$, the uncertainty is the highest. When $q \sim U(0, L)$, the scaled dot-product self-attention becomes the mean value of $V$, which is redundant for residential input. Therefore, we keep the distribution, $p$, as far as possible from the distribution, $q$. Then, we use the Kullback–Leibler (KL) divergence to measure the distance between the distributions:

$$KL(q \| p) = \mathcal{H}(q, p) - \mathcal{H}(q) = \sum q \log \frac{q}{p}, \tag{11}$$

where $\mathcal{H}(q, p)$ is the cross entropy. We substitute $q(\mathbf{k}_j | \mathbf{q}_i) = 1/L$ and $p(\mathbf{k}_j | \mathbf{q}_i) = f(\mathbf{q}_i, \mathbf{k}_j) / \sum_k f(\mathbf{q}_i, \mathbf{k}_k)$ into Eq. (11) to get

$$KL(q \| p) = \sum_{j=1}^{L} \frac{1}{L} \log \frac{\frac{1}{L}}{\frac{f(\mathbf{q}_i, \mathbf{k}_j)}{\sum_k f(\mathbf{q}_i, \mathbf{k}_k)}} = \log \sum_{j=1}^{L} \exp\left(\frac{\mathbf{q}_i \mathbf{k}_j^\top}{\sqrt{d}}\right) - \frac{1}{L} \sum_{j=1}^{L} \frac{\mathbf{q}_i \mathbf{k}_j^\top}{\sqrt{d}} - \log L, \tag{12}$$

where $KL(q \| p)$ is bounded. Its lower bound is set by letting $h(x) = -\log x$; then, the function is strictly convex, and $\sum q(\mathbf{k}_j | \mathbf{q}_i) = 1$. According to the Jensen inequality,

$$KL(q \| p) = \sum_{j=1}^{L} \frac{1}{L} \log \frac{\frac{1}{L}}{\frac{f(\mathbf{q}_i, \mathbf{k}_j)}{\sum_k f(\mathbf{q}_i, \mathbf{k}_k)}} \geq -\log \sum_{j=1}^{L} \frac{f(\mathbf{q}_i, \mathbf{k}_j)}{\sum_k f(\mathbf{q}_i, \mathbf{k}_k)} = -\ln 1 = 0, \tag{13}$$

and its upper bound is

$$KL(q \| p) = \log \sum_{j=1}^{L} \exp\left(\frac{\mathbf{q}_i \mathbf{k}_j^\top}{\sqrt{d}}\right) - \frac{1}{L} \sum_{j=1}^{L} \frac{\mathbf{q}_i \mathbf{k}_j^\top}{\sqrt{d}} - \log L \leq \log \sum_{j=1}^{L} \exp \max_j \left(\frac{\mathbf{q}_i \mathbf{k}_j^\top}{\sqrt{d}}\right) - \frac{1}{L} \sum_{j=1}^{L} \frac{\mathbf{q}_i \mathbf{k}_j^\top}{\sqrt{d}} - \log L = \max_j \left(\frac{\mathbf{q}_i \mathbf{k}_j^\top}{\sqrt{d}}\right) - \frac{1}{L} \sum_{j=1}^{L} \frac{\mathbf{q}_i \mathbf{k}_j^\top}{\sqrt{d}}. \tag{14}$$

Therefore, we choose the largest *Nd* sequence steps according to KL divergence to replace the original full self-attention. However, the existence of exponential function in Eq. (12) leads to an unstable numerical overflow in the calculation of KL divergence. Thus, we use the approximate value instead:

$$M = \max_j \left( \frac{q_i k_j^\top}{\sqrt{d}} \right) - \frac{1}{L} \sum_{j=1}^{L} \frac{q_i k_j^\top}{\sqrt{d}}. \tag{15}$$

The substituted $M$ can be understood as the difference between the dominant term and the mean. If the difference between them is small, then it can be ignored. During the continuous training process, the possibility is higher that this data is noise.

We calculate the $M$ value of all $q_i$, $i = 1, 2, …, L$ and choose $\boldsymbol{Q}_s \in \mathbb{R}^{Nd \times d}$ instead of $\boldsymbol{Q}$ as the query in the sparse attention. Therefore, the full attention of Eq. (3) becomes

$$A(\boldsymbol{Q}_s, \boldsymbol{K}, \boldsymbol{V}) = \text{Softmax}\left( \frac{\boldsymbol{Q}_s \boldsymbol{K}^\top}{\sqrt{d}} \right) \boldsymbol{V}. \tag{16}$$

We then concatenate the outputs of the multi-head attention and add them to the inputs after normalization:

$$\boldsymbol{X}_A^s = \phi \left( \boldsymbol{X}_A + A(\boldsymbol{Q}_s, \boldsymbol{K}, \boldsymbol{V}) \right), \tag{17}$$

where $\phi$ is the normalization function. Then, the feedforward layer is adopted as

$$\boldsymbol{X}_A^{out} = \phi \left( \boldsymbol{X}_A + \text{ReLU}(\boldsymbol{X}_A^s \boldsymbol{W}_{A1} + \boldsymbol{b}_1) \boldsymbol{W}_{A2} + \boldsymbol{b}_2 \right), \tag{18}$$

where $\boldsymbol{W}_{A1} \in \mathbb{R}^{d_{model} \times 4d_{model}}$, $\boldsymbol{b}_{A1} \in \mathbb{R}^{4d_{model}}$, $\boldsymbol{W}_{A2} \in \mathbb{R}^{4d_{model} \times d_{model}}$, $\boldsymbol{b}_{A2} \in \mathbb{R}^{d_{model}}$, representing is the weight of linear layers.

After the sparse self-attention, to improve the influence of useful features again, we use the convolution-pooling unit. A shortcut is not used here to reduce further noise propagation. Max-pooling selects more significant convolution features to better suppress the influence of noise. After $Nx$ sparse attention Conv1d blocks, the dynamic and kinematic features of the agent are encoded.

*3.2.1.2 Lane vector encoding*

LaneGCN is adopted to encode the lane vectors [13]. Owing to the relatively high speeds of agents traveling long distances in a lane, convoluting only one predecessor or successor lane vector is insufficient. To capture the long-range dependencies, a dilated LaneConv for LaneGCN is used to focus more predecessors and successors at a multi-scale level:

$$\boldsymbol{M}_L^{out} = \boldsymbol{M}_f \boldsymbol{W}_f + \sum_{i \in \{r,l,m,o\}} \boldsymbol{M}_i \boldsymbol{M}_f \boldsymbol{W}_i + \sum_{c=1}^{C} \left( \boldsymbol{M}_p^{k_c} \boldsymbol{M}_f \boldsymbol{W}_p + \boldsymbol{M}_s^{k_c} \boldsymbol{M}_f \boldsymbol{W}_s \right), \tag{19}$$

where $\boldsymbol{W}_.$ is the weight of LaneConv, and $k_c$ is the $c$th dilation size. Normalization and rectified linear units are adopted after each LaneConv and linear layer. After $Ng$ LaneGCN blocks, the static lane vector features, $\boldsymbol{X}_{SL}^{out}$, are finally encoded.

The static lane vector features, $\boldsymbol{X}_{SL}^{out}$, describe the topological structure of the lane of interest, but it lacks the dynamic scene information of the agent on the scene. Dynamic scene information refers to the additional properties generated when an agent uses the scene. There are also differences in the maneuvers made by the agent in response to road blockages, non-right-of-lane-owner intrusion scenarios, and ordinary scenarios. Therefore, dynamic lane vector features are very important for agent trajectory prediction and decision-making.

To extract the dynamic lane vector features, we use the static lane line features, $X_{SL}^{out}$, as the basis and integrate the information of each agent into the lane vectors through the feature selection sparse attention mechanism. Feature selection sparse attention is described in the next section. Then, the information of each lane line is transmitted again through LaneGCN so that the lane line has traffic flow and agent lane occupancy information. After $Ng$ LaneGCN blocks, the dynamic lane vector features, $X_{DL}^{out}$, are finally encoded.

*3.2.2 Feature selection network*

After encoding, the encoder obtains three cues: agent kinematics and static and dynamic lane vectors. When these cues are fused, the irrelevant features make neural network learning more difficult. Unlike the original input of the agent lacking prior knowledge, the agent often only pays attention to the lane and other agents within a certain range. Based on this prior knowledge, we can purposefully select scene elements requiring attention. Therefore, we propose a feature selection network that combines multiple cues.

The input of feature selection sparse attention is of the *base* and *context* types; that is, base pays attention to useful features in context under partial prior knowledge. First, the tensor-of-interest selector is used to select the set of interesting features according to the Euclidean distance between base and context:

$$S_{tp} = \{(x_i, x_j, d_{ij}) \mid d_{ij} = d_i - d_j, \|d_{ij}\|_2 < \varepsilon_{th}, i = \{1, 2, ..., |X_{base}|\}, j = \{1, 2, ..., |X_{ctx}|\}, x_i \in X_{base}, x_j \in X_{ctx}\}, \tag{20}$$

where $d_i$, $d_j$ is the coordinate of $x_i$, $x_j$, respectively, in the local coordinate system, and $\varepsilon_{th}$ is the threshold of the ROI. Then the first, second, and third items of the tuples in the set are fed to a linear layer and recorded as $\bar{X}_{base}$, $\bar{X}_{ctx}$, and $\bar{D}_{bc}$, respectively. We connect the context with the distance information, $\bar{D}_{bc}$, and combine it with base through the linear layer, activating it with the exponential linear unit (ELU) after normalization:

$$\bar{X}_{bc} = \text{ELU}(\phi(\bar{X}_{base}W_b + b_b + \text{concat}(\bar{X}_{ctx}, \bar{D}_{bc})W_c)), \tag{21}$$

where $W_b$, $W_c$ is the linear weight, and $b_b$ is the bias. When $\bar{X}_{base}W_b + b_b + \text{concat}(\bar{X}_{ctx}, \bar{D}_{bc})W_c \gg 0$, ELU activation acts as an identity function; when $\bar{X}_{base}W_b + b_b + \text{concat}(\bar{X}_{ctx}, \bar{D}_{bc})W_c \ll 0$, it generates a constant output, resulting in linear layer behavior. Therefore, we adopt the gate selection mechanism to select the useful features in the context:

$$\gamma = \bar{X}_{bc}W_\gamma + b_\gamma, \tag{22}$$

$$\bar{X}_\gamma = \sigma(W_\sigma \gamma + b_\sigma) \odot (W_g \gamma + b_g), \tag{23}$$

$$\bar{X}_g = \text{ReLU}(\phi(\bar{X}_{bc} + \bar{X}_\gamma)), \tag{24}$$

where $W_\gamma$, $W_\sigma$, and $W_g$ are the linear weights, $b_\gamma$, $b_\sigma$, and $b_g$ are the biases, and $\odot$ is the Hadamard product. We then add and normalize the selected features to the original input:

$$X_{b,c}^{att} = \phi(X_{base} \oplus \bar{X}_g), \tag{25}$$

where $\oplus$ indicates an index-add, and $X_{b,c}^{att}$ is recorded as the output of feature selection sparse attention. Then, after the feedforward layer (Eq. (18)), the fused output, $X_{b,c}^{out}$, is the base feature fused to the context feature.

Through the aforementioned operations, the base selects features in the context of interest for feature fusion, which reduces the learning difficulty and enables the organic fusion of multiple cues. In the feature selection network, the base of the lane-feature selection sparse attention refers to agents, and the context refers to lane vectors. This attention is used to establish the interaction between the agent and the surrounding environment. The base of the agent feature selection sparse attention is the target agent, and the context is other agents. This attention is used to establish the interaction between different agents. Additionally, the outputs of the above two interaction feature selection sparse attentions, $X_{b,c}^{att}$, are taken out for subsequent agent-maneuver decision-making.

After the interaction is established, to fully use the posterior knowledge of the lane decision, a lane-scaled dot-product attention is adopted. Notably, as there may be multiple or no lane lines in the attention layer, a sigmoid function is used instead of Eq. (3) for activation:

$$A_S(Q, K, V) = \text{Sigmoid}\left(\frac{QK^\top}{\sqrt{d}}\right)V, \tag{26}$$

where the attention output, $X_{lane}^{att} = \text{Sigmoid}(QK^\top/\sqrt{d})$, of the sigmoid is taken out for subsequent lane decisions. After the attention, the agent features, $X_{fuse}^{out}$, are combined with multi-cues are used for decoding.

*3.2.3 Decoder*

The decoder consists of two parts: behavior decisions and trajectory predictions. In the behavior decision branch, the lane line attention weight is the lane decision probability, $\hat{P}_{lane}^{dec} = X_{lane}^{att}$, where $\hat{P}_{lane}^{dec} \in [0,1]^{(Ns+1) \times Nl}$. The agent maneuver decision is concatenated by multiple $X_{b,c}^{att}$ and is output through the linear layer to obtain the agent maneuver decision probability, $\hat{P}_{mane}^{dec} \in [0,1]^{(Ns+1) \times 1}$.

The trajectory prediction module takes $X_{fuse}^{out}$ as the input of the linear ResNet block and decodes $\hat{A}_{traj} \in \mathbb{R}^{(Ns+1) \times K \times t_{fut} \times 2}$. To improve the network's ability to explore the multimodality of agents, feature selection attention is adopted so that the feature selection sparse attention of the tensor-of-interest selector is removed. The base is $X_{fuse}^{out}$, and the context is the predicted value, $\hat{A}_{traj}^{t_{fut}} \in \mathbb{R}^{(Ns+1) \times K \times 2}$, at the last moment in $\hat{A}_{traj}$. Finally, the probability, $\hat{P}_{traj}^{pred} \in [0,1]^{(Ns+1) \times K \times 1}$, of each predicted trajectory is output through the SoftMax layer.

## 3.3 Loss function

The loss function consists of trajectory prediction and behavior decision losses. The trajectory prediction task predicts $K$ trajectories and their corresponding probabilities. For any agent, let $k^* = \arg\max(\boldsymbol{P}_{traj}^{pred})$ to locate the trajectory with the greatest probability. To improve the ability of the model to explore the multimodality of an agent, the max-margin loss is adopted:

$$J_M = \frac{\sum_{ns=1}^{Ns+1} \sum_{k \neq k^*} \max(0, \hat{p}_{ns}^k - \hat{p}_{ns}^{k^*} + \varepsilon_M)}{(Ns+1)(K-1)}, \tag{26}$$

where $\hat{p}_{ns}^k$ is the probability of the $k$th modality of the $ns$th agent, and $\varepsilon_M$ is the margin.

To improve the confidence of the $k^*$th trajectory, a pseudo label, $\tilde{p}^{k^*}$, is set. That is, $k^*$ assigns a positive label, and $k \neq k^*$ are assigned negative labels. Then cross-entropy loss is adopted:

$$J_C^{traj} = -\frac{\sum_{ns=1}^{Ns+1} \sum_{c=1}^{K} \tilde{p}_{ns}^{k^*} \log \hat{p}_{ns}^c}{Ns+1}. \tag{27}$$

Additionally, the error between predicted and the real trajectories is calculated using the smooth $L1$ loss:

$$J_S = \frac{1}{Ns+1} \sum_{ns=1}^{Ns+1} \sum_{j=1}^{t_{fut}} \begin{cases} \frac{1}{2} \left\| \hat{a}_{j,k^*}^{ns} - a_j^{ns} \right\|_2^2, & \text{if } \left\| \hat{a}_{j,k^*}^{ns} - a_j^{ns} \right\| < 1 \\ \left\| \hat{a}_{j,k^*}^{ns} - a_j^{ns} \right\| - \frac{1}{2}, & \text{otherwise,} \end{cases} \tag{28}$$

where $a_j^{ns}$ is the ground truth of the $ns$th agent at time $j$.

Behavior decision tasks are classified tasks, and cross-entropy loss is adopted:

$$J_C^{lane} = -\frac{\sum_{nl=1}^{Nl} \sum_{ns=1}^{Ns+1} (D_l \log \hat{D}_l + (1-D_l) \log(1-\hat{D}_l))}{Nl(Ns+1)}, \tag{29}$$

$$J_C^{agent} = -\frac{\sum_{ns=1}^{Ns+1} \sum_{c=1}^{6} D_m \log \hat{D}_m}{(Ns+1)}. \tag{30}$$

The final total loss is the sum of the above losses:

$$J = J_M + J_C^{traj} + J_S + J_C^{lane} + J_C^{agent}. \tag{31}$$

## 4 EXPERIMENTS AND EVALUATIONS

### 4.1 Experimental setting

The Argoverse public motion dataset used in this study contains agent information of cars, bicycles, and pedestrians, as well as HD-map information collected on urban roads in Pittsburgh and Miami in the United States. All data were collected by vehicles equipped with LiDAR and cameras at a frequency of 10 Hz. The dataset was split into training, validation, and testing sets with 205,942, 39,472, and 78,143 sequences, respectively. The future trajectory of the test set was not provided, and there was no geographic overlap between sets. A 5-s period was selected to describe the trajectory of each vehicle: 2 s were used as historical input, and 3 s were used as the trajectory to be predicted.

All experiments were performed on an Intel Core(R) i9-10920x CPU @ 3.50 GHz (Turbo 4.60 GHz), NVIDIA GeForce(R) RTX 3090 24-GB GPU with 32-GB of RAM running the Ubuntu 16.04 LTS edition. All program tasks were conducted on Python 3.7, and the deep learning framework was based on PyTorch.

*4.2 Training setting and evaluation metrics*

As described, the length of the input feature sequence of agents was $t_{hst}$=20. The 180 trajectory points of $t_{fut}$ = 30 and $K = 6$ modals in the future were predicted, and the agent lane and maneuver decisions were output. The feature dimension, $d_{model}$, of the model was set to 128. In the sparse multi-head attention, $h = 4$ heads were selected. The largest $Nd = 0.75L$ sequence steps were used according to KL divergence to replace the original full self-attention. The sparse multi-head attention block, $Nx$, the LaneGCN block, $Ng$, and the feature selection sparse attention block, $Na$, were set to three, four, and two, respectively. $\varepsilon_M$ was set to 0.2. The Adam optimizer was employed using a learning rate of $10^{-3}$ and a mini-batch size of 32. Because the ground truth of the test set was not provided in the dataset, the trajectory prediction metrics of the model came from the test set, and other results came from the verification set.

For prediction performance evaluation, average displacement error (ADE) and final displacement error (FDE) were adopted. ADE is the average L2 distance between the prediction trajectory and the ground truth, and FDE is the L2 distance between endpoint of the prediction trajectory and the ground truth. In this study, minimum ADE (minADE) and minimum FDE (minFDE) at $K = 1$ and $K = 6$, respectively, were used as evaluation metrics to obtain the performance of the best predicted trajectory under single- and multi-modal cases. For decision performance evaluation, accuracy, precision, recall, and F1-score were used as basic model evaluation indicators. Furthermore, to better demonstrate the performance of the model in specific cases for maneuvering decisions, additional statistics were taken on *yield* decisions. For lane decisions, case-level recall was evaluated when at least one or more correct lane lines were identified.

*4.3 Results and comparisons*

To validate the proposed model, its results were compared to those of models proposed by other excellent studies conducted in recent years:

- Nearest-neighbor (NN) regression (Argoverse Baseline) [36]: The baseline builds on NN and prunes the number of predicted trajectories based on how often they exit the drivable area.

- Target-driven trajectory (TNT) [28]: A prediction framework that contains three training stages.
- DenseTNT [37]: An anchor-free, end-to-end trajectory prediction model that directly outputs a set of trajectories from dense goal candidates.
- VectorNet [12]: A hierarchical graph neural network that first exploits the spatial locality of individual road components represented by vectors, then it models the high-order interactions among all components.
- LaneGCN [13]: A lane-graph CNN that captures the complex topology and long-range dependencies of the lane graph.
- Temporal point-cloud network (TPCN) [30]: Extends ideas from point-cloud learning with dynamic temporal learning to capture both spatial and temporal information by splitting trajectory prediction into those same dimensions.
- Multi-modal transformer [38]: A neural prediction framework based on the transformer structure that models the relationship among interacting agents and extract the attention of the target agent on map waypoints.
- Heatmap output for future motion estimation (HOME) [39]: A framework that tackles the motion forecasting problem with an image output representing the probability distribution of the agent's future location.
- Scene transformer [40]: A model using masking strategy as the query, enabling one to invoke a single model to predict agent behavior in many ways.
- HT: The model proposed in this study.

**Table 3** shows minADE and minFDE at $K$=1 and $K$=6, respectively. We see that the performance of the HT was slightly worse than that of scene transformer on $K$=6 minADE, and the proposed method performed best comprehensively with multiple performance metrics. LaneGCN and VectorNet are based on vector representation. Among them, VectorNet simply connects the features through a graph neural network. LaneGCN uses a feature pyramid network to encode the historical trajectory of agents, fuse lane features with agent features through the attention mechanism, and output the trajectory of agents alongside the predicted endpoint attention mechanism. HT is also based on vectorized input representation, and it showed great improvements in all evaluation metrics. Compared with the above two methods, HT learned the relationship between vectors more deeply, making learning, dissemination, and feature and information fusion between the agent and the road more efficient and comprehensive. Multi-modal and scene transformers are like vanilla methods. The multi-modal transformer trains stacked transformer blocks using a tailored region-based training strategy. The scene transformer realizes multi-input fusion by stacking transformer blocks across times, agents, and road graphs. The method based on the transformer performed better than other methods, indicating that the transformer did well with multi-source information extraction and fusion. Compared with the aforementioned two methods, decision-making auxiliary regularization was added to HT, which reduced the difficulty of model training.

**Table 3**
Metrics of each model in the 3 s prediction horizon on the Argoverse test set

| Model | K=6 | | K=1 | |
|---|---|---|---|---|
| | minADE | minFDE | minADE | minFDE |
| NN (baseline) [36] | 1.713 | 3.287 | 3.455 | 7.883 |
| VectorNet [12] | 3.110 | 6.723 | 3.110 | 6.723 |
| TNT [28] | 0.910 | 1.446 | 2.174 | 4.959 |
| HOME [39] | 0.890 | 1.292 | 1.699 | 3.681 |
| DenseTNT [37] | 0.882 | 1.282 | 1.679 | 3.632 |
| LaneGCN [13] | 0.878 | 1.355 | 1.702 | 3.764 |
| Multi-modal Transformer [38] | 0.837 | 1.291 | 1.735 | 3.901 |
| TPCN [30] | 0.815 | 1.244 | 1.575 | 3.487 |
| Scene Transformer [40] | **0.803** | 1.232 | 1.811 | 4.055 |
| HT (proposed) | 0.811 | **1.223** | **1.571** | **3.435** |

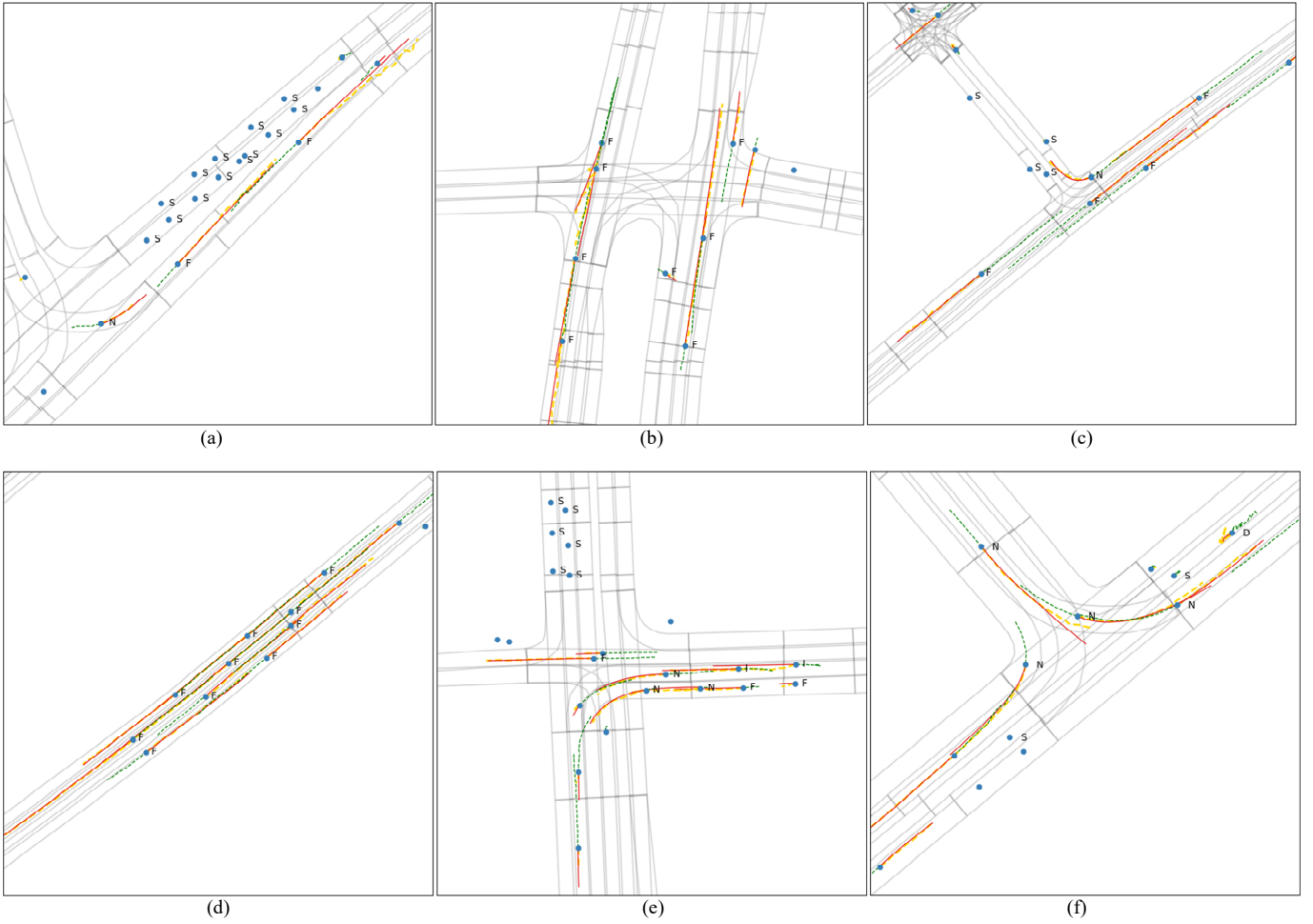

**Figure 3.** Prediction instances from the bird's-eye view of Argoverse validation dataset. For simplicity, only the $k^*$-th trajectory is displayed. The blue dot is the agent, the green dotted line is the historical trajectory of the agent, the red solid line is the predicted trajectory, and the yellow dotted line is the ground truth. The characters next to the blue dots represent the behavior decision classification results of the agent.

**Figure 3** shows prediction instances from the bird's-eye view of the Argoverse validation dataset. It can be seen that the proposed model output accurate prediction results for all agents in and out of junction. In **Figure 3** (e), there were multiple agents entering the junction and choosing to turn left. The model in this case output accurate prediction results, and "N" indicates that the agent's decision was a lane-change. In the leftmost lane, the agent behind the lane-changing agent had not yet entered the junction, so the decision information output, "F," indicated following the agent in front. The decision-making output of the agent in the second left lane on the left was "I," indicating that the agent preferred not to change lanes. **Figure 3** (f) shows a T-shaped junction scenario. Although the agent at the upper left in the figure did not enter the junction, the decision information output by the model was "N," indicating that it was about to change lanes. This shows that the model captured the interaction between the agent and the surrounding environment and output accurate and reasonable trajectory predictions and behavior decision-making results.

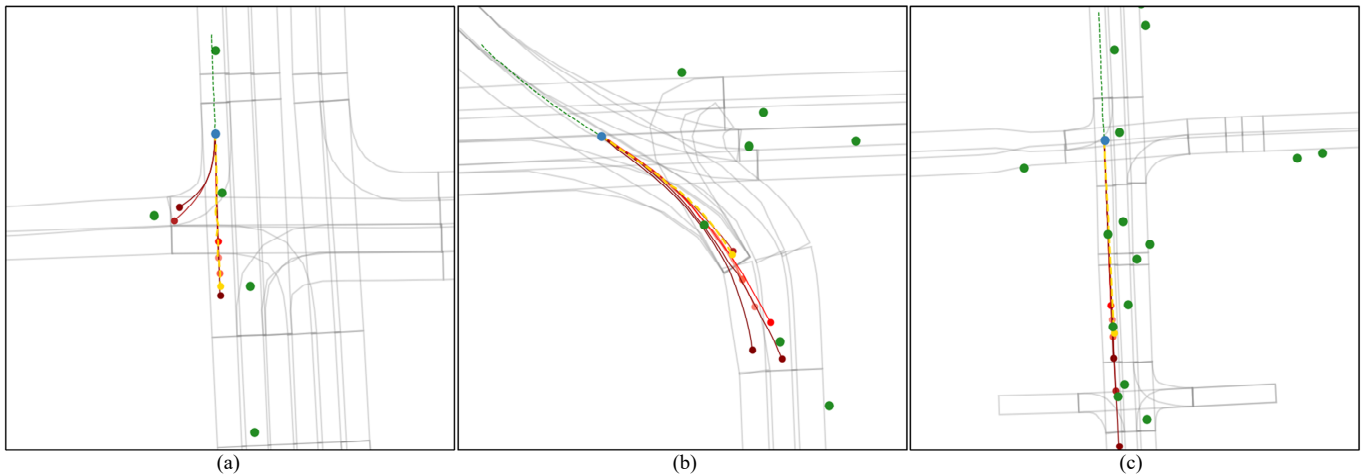

(a)            (b)            (c)

**Figure 4.** Multi-modal prediction instances from the bird's-eye view of Argoverse validation dataset. For simplicity, prediction results are shown for only one agent (blue dots). The green dots are other agents, the yellow dotted line is the ground truth, and the other six lanes are the predicted trajectories.

**Figure 4** shows multi-modal prediction instances from the bird's-eye view of the Argoverse validation dataset. **Figure 4** (a) and (b) show the possible future trajectories of the agent according to the current road topology and the interactions of surrounding agents. For example, in **Figure 4** (a), the agent was located in the rightmost lane, and the model output not only four straight-ahead prediction trajectories, but it also supplied two right-turn prediction trajectories. When the agent in **Figure 4** (b) drove out of the junction, the model provided a variety of possible trajectories, and the end points were located in two different lanes. Such prediction results reflect the model's inferencing power of agent motion under existing conditions, and it fully considered the results of the interaction of road topologies and agents.

**Table 4**
Classification performance of each maneuver decision for the Argoverse validation dataset

| Item | | Precision (%) | Recall (%) | F1-Score (%) |
|---|---|---|---|---|
| **Maneuver Decision** | S | 88.07 | 80.76 | 84.26 |
| | N | 62.77 | 77.02 | 69.17 |
| | D | 55.96 | 71.22 | 62.67 |
| | F | 89.92 | 78.09 | 83.59 |
| | I | 63.88 | 82.25 | 71.91 |
| *Yield* **Decision** | | 86.05 | 90.01 | 87.99 |
| **Lane Decision** | | 82.76 | 68.55 | 74.99 |
| **Maneuver Decision Accuracy (%)** | | | 78.57 | |
| **Lane Decision Accuracy (%)** | | | 99.62 | |
| **Case-Level Lane Decision Recall (%)** | | | 89.98 | |

**Table 4** shows the results of maneuver decision classification on the Argoverse validation set. Generally, maneuver decisions "S" and "F" showed the best classification performance, and "N", "D," and "I" showed general classification performance. Notably, decisions "S", "N", and "D" all belonged to the yield class, in which classification errors are prone to occur. Therefore, it can be seen that the model had good classification performance (F1-score: 87.99%) when the metrics of the yield class were counted. Similarly, lane decision performance can only reflect model performance, making it difficult to attribute to specific cases. Owing to the class imbalance between positive and negative lane decision classes (~1:100), decision accuracy had little significance. Therefore, by calculating the recall of lane decisions at the case-level, it can be seen that the model output lane decision results more accurately (case-level lane decision recall: 89.98%).

*4.4 Decision posterior knowledge attention visualization*

In the feature selection network, the sigmoid activation function was used for the lane-scaled dot-product attention. Therefore, each agent corresponded to an attention weight for all lane lines with a value range of [0,1]. The value indicates the attention of the model to the target lane segment at the current time. To show the best interpretable results consistent with human intuition, **Figure 5** shows the decision posterior knowledge attention results. From these, the following conclusions can be drawn:

1) **Figure 5** (a) and (b) are lane-changing cases in the out-of-junction scenario of the target agent. In **Figure 5** (a), the target agent was about to change lanes to surpass the agent in front. The lane in front of the target agent and the one in the front-left of the target agent had high attention weights. In **Figure 5** (b), the target agent was about to complete a lane change, and its lane attention had a high weight directly only to the target lane line. This is similar to the driver's attention focusing on the lane line during a lane change. In such a case, the driver will always observe the current and target lanes before lane changing. Then, the driver decides to change lanes. When the target agent crosses the lane line, the attention of the original lane decreases gradually, and the attention is finally completely focused on the target lane.

2) **Figure 5** (c), (d), (e), and (f) show the scenes of the agent in the junction. **Figure 5** (c), (d), and (e) show agents about to enter the junction, which are divided into combinations of four and two modals based on the predicted trajectories, focusing on different lane lines. Notably, the model captured the attention of the target agent across multiple lane lines, and it output two feasible lane decisions: straight and turning. Notably, the output of the model was not the final decision information. The final decision must be limited by global routing, traffic regulations, and other information, which are not discussed in this study. **Figure 5** (f) shows that the agent determined the proper lane line and was about to leave the junction, focusing on the target lane, which is similar to the lane change maneuver.

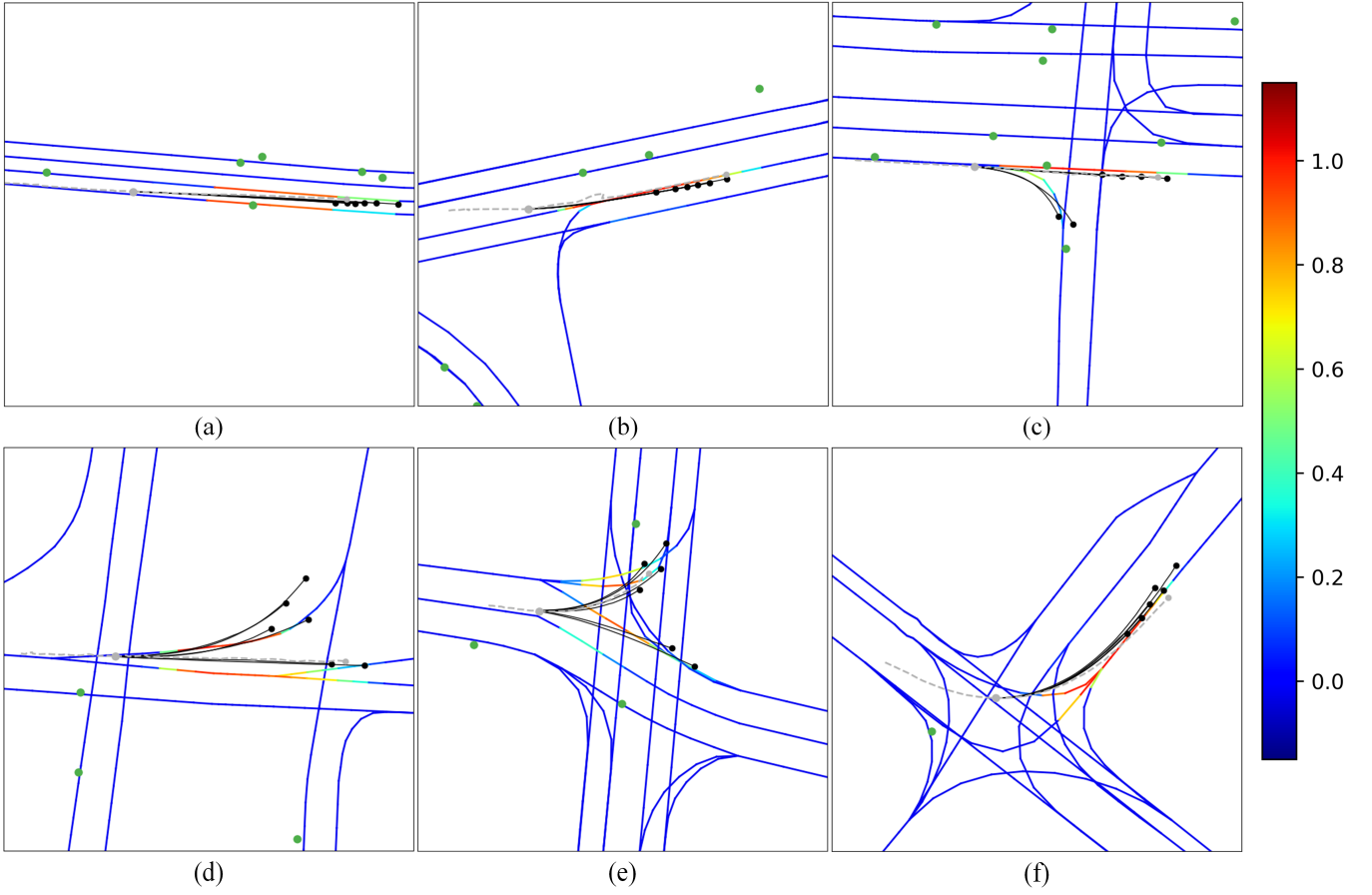

**Figure 5**. Schematic of the decision posterior knowledge attention visualization. The gray dot is the target agent, the gray dotted line is the ground truth trajectory, the black solid line is the predicted trajectory of $K$=6 modals, and the green dot is the surrounding agent. The other solid lines are the lane center lines. When the line is red, the lane line attention is the largest, and when it is blue, the lane line attention is the smallest.

*4.5 Perceptual noise robustness experiment*

As described in **Section 2**, the input data used for trajectory prediction are vulnerable to noise, resulting in data-dependent aleatoric uncertainty, which lead to worsening trajectory prediction and behavioral decision-making performances. Therefore, we proposed a sparse multi-head attention for agent trajectory encoding. To verify its robustness to noise, we designed two experimental scenarios. The first simulated perceptual loss, in which one frame of data in the input data was replaced by the probability of $p_o$, has a value of zero. The second scenario simulated perceptual noise, in which Gaussian noise was added to one frame of the input data with probability $p_n$, where $p_n \sim \mathcal{N}(0, v/100)$ and $v$ is the velocity of the agent. To better compare the vanilla and sparse multi-head attentions, the experimental comparison model involved replacement of the sparse multi-head attention with the vanilla multi-head attention.

**Table 5** shows the mean and variance values of the prediction results under the influence of different noise. The mean shows the accuracy of prediction, and the variance shows its stability. The smaller the mean and variance, the more accurate and stable the prediction result of the model. For ease of illustration, **Figure 6** shows the trajectory prediction performance degradation of the two comparison models under the influence of noise. According to **Table 5** and **Figure 6**, the following conclusions can be drawn:

**Table 5**
Comparison of mean and variance values of prediction results under the influence of different noise

| Method | Item | | Mean | | | | Variance | | | |
| --- | --- | --- | --- | --- | --- | --- | --- | --- | --- | --- |
| | | | K=1 | | K=6 | | K=1 | | K=6 | |
| | | | minADE | minFDE | minADE | minFDE | minADE | minFDE | minADE | minFDE |
| Sparse Multi-head Attention | Baseline | | 1.262 | 2.722 | 0.636 | 0.957 | 3.488 | 10.117 | 0.630 | 1.301 |
| | Case1: $p_o$ | 0.01 | 1.289 | 2.775 | 0.644 | 0.967 | 3.591 | 10.361 | 0.637 | 1.301 |
| | | 0.03 | 1.357 | 2.911 | 0.661 | 0.996 | 3.953 | 11.262 | 0.671 | 1.389 |
| | | 0.05 | 1.431 | 3.053 | 0.678 | 1.024 | 4.248 | 11.962 | 0.687 | 1.428 |
| | | 0.08 | 1.556 | 3.305 | 0.705 | 1.064 | 4.837 | 13.281 | 0.743 | 1.502 |
| | Case2: $p_n$ | 0.01 | 1.302 | 2.798 | 0.647 | 0.971 | 3.688 | 10.626 | 0.643 | 1.319 |
| | | 0.03 | 1.391 | 2.965 | 0.668 | 1.001 | 4.092 | 11.565 | 0.683 | 1.408 |
| | | 0.05 | 1.501 | 3.181 | 0.691 | 1.038 | 4.652 | 12.904 | 0.726 | 1.483 |
| | | 0.08 | 1.706 | 3.572 | 0.733 | 1.104 | 5.665 | 14.628 | 0.811 | 1.646 |
| Vanilla Multi-head Attention | Baseline | | 1.429 | 3.020 | 0.716 | 1.011 | 3.939 | 11.228 | 0.721 | 1.485 |
| | Case1: $p_o$ | 0.01 | 1.474 | 3.104 | 0.727 | 1.028 | 4.111 | 11.621 | 0.737 | 1.512 |
| | | 0.03 | 1.573 | 3.293 | 0.749 | 1.060 | 4.604 | 12.792 | 0.782 | 1.614 |
| | | 0.05 | 1.667 | 3.467 | 0.770 | 1.097 | 5.027 | 13.658 | 0.818 | 1.675 |
| | | 0.08 | 1.811 | 3.746 | 0.802 | 1.148 | 5.715 | 15.202 | 0.927 | 1.956 |
| | Case2: $p_n$ | 0.01 | 1.486 | 3.126 | 0.729 | 1.030 | 4.197 | 11.831 | 0.746 | 1.539 |
| | | 0.03 | 1.601 | 3.345 | 0.756 | 1.073 | 4.770 | 13.143 | 0.819 | 1.699 |
| | | 0.05 | 1.739 | 3.607 | 0.788 | 1.124 | 5.462 | 14.670 | 0.896 | 1.866 |
| | | 0.08 | 1.933 | 3.965 | 0.837 | 1.203 | 6.338 | 16.491 | 1.043 | 2.240 |

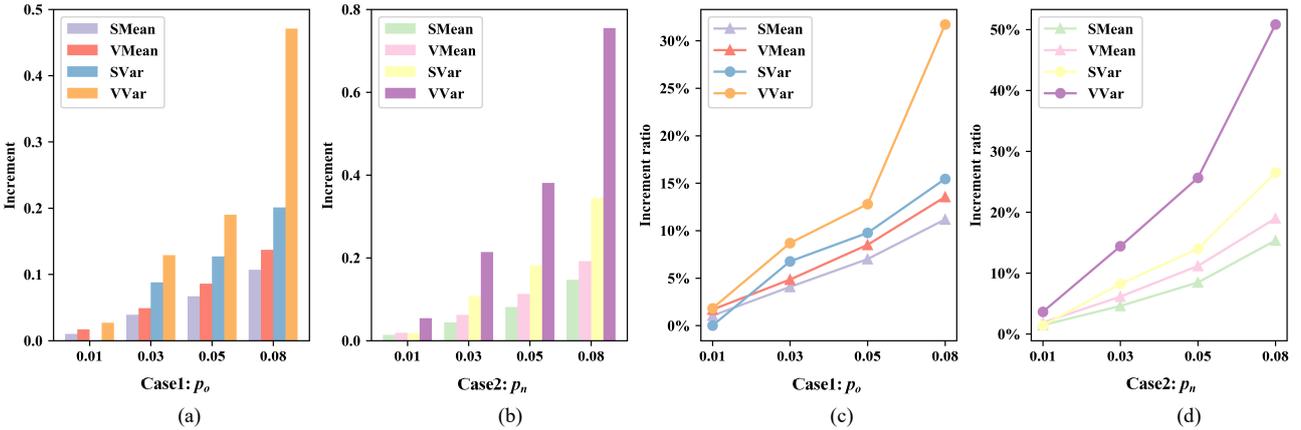

**Figure 6**. Schematic of trajectory prediction performance degradations of the two comparison models under the influence of noise. In the figure, minFDE with modal $K = 6$ is selected as the example, and the results of the baseline model are compared. Figures (a) and (b) show the minFDE increment, and (c) and (d) show the ratio of minFDE increment.

1) Because the original training data contained a lot of noise, the prediction error of sparse multi-head attention was smaller than that of the vanilla multi-head attention in the baseline model, and the prediction result was more stable. Under the influence of noise, the performance of both contrasting models decreased, and that of the minFDE increased with noise.

2) The sparse multi-head attention was significantly lower than the vanilla multi-head attention in terms of increment and increment ratio minFDE metrics. Additionally, when the noise was large, the mean increment ratio and sparse multi-head attention variance increment ratio of the two methods were similar, but the vanilla multi-head attention variance had a large increase. This is because there were a smaller number of data having large prediction errors in the vanilla multi-head attention model. This illustrates that sparse multi-head attention is more robust to perceptual noise.

3) By contrast, perceptual noise may be more influential than perceptual loss. This tells us that if there is obviously a large noise in the input data of one frame, the trajectory can be smoothed through preprocessing; otherwise, the frame can be directly eliminated to reduce the impact of noise.

*4.6 Comparison and ablation experiment*

To verify the performance of each module of the proposed model, the following comparison and ablation experiments were conducted:

- No feature selection: The feature selection network was removed, and only lane-scaled dot-product attention was retained.
- Full feature selection: The tensor-of-interest selector in the feature selection sparse attention was removed.
- LSTM Encoder: The sparse multi-head attention in agent trajectory encoding was replaced with LSTM.
- Vanilla Attention: The sparse multi-head attention in agent trajectory encoding was replaced with the vanilla multi-head attention.
- No decision: All decision branches were replaced.
- No $X_{lane}^{att}$: Lane decision branches were replaced.
- No $X_{b,c}^{att}$: Agent behavior decision branches were replaced.
- HT: Full model proposed in this study.

**Table 6** shows the metrics of ablation in the 3-s prediction horizon using the Argoverse validation set. The following inferences were obtained based on these results:

**Table 6**
Metrics of the ablation model in the 3 s prediction horizon for the argoverse validation set

|  | Metrics | No feature selection | Full feature selection | LSTM Encoder | Vanilla Attention | No decision | No $X_{lane}^{att}$ | No $X_{b,c}^{att}$ | HT (proposed) |
|---|---|---|---|---|---|---|---|---|---|
| $K=1$ | minADE | 1.623 | 1.377 | 1.435 | 1.429 | 1.250 | **1.236** | 1.289 | 1.262 |
|  | minFDE | 3.622 | 2.977 | 3.061 | 3.020 | 2.691 | **2.669** | 2.727 | 2.722 |
| $K=6$ | minADE | 0.762 | 0.732 | 0.721 | 0.716 | 0.668 | 0.658 | 0.659 | **0.636** |
|  | minFDE | 1.193 | 1.089 | 1.027 | 1.011 | 0.990 | 0.976 | 0.972 | **0.957** |

1) No feature selection vs. full feature selection: Without feature selection, only the interactions between agents and between agents and lane lines were captured in the final decoding part; hence, performance dropped the most. If full attention feature selection is used, the performance is improved, but complex scenes may introduce useless information, which affects trajectory prediction performance. Therefore, guided by some prior knowledge, feature selection sparse attention can learn more accurate features.

2) LSTM encoder, vanilla attention, and the proposed model: The performance of the model similar to the vanilla transformer was better than that of LSTM, indicating that the multi-head attention can better extract various features across time steps.

The comparison between vanilla attention and the proposed model was not repeated.

3) No decision, no $X_{lane}^{att}$, no $X_{b,c}^{att}$, and the proposed model: Prediction performance of the proposed model was not optimal when $K$ = 1. When the lane attention was removed and only the agent behavior decisions were kept, the model predicted the best performance when $K$ = 1. When only the lane decision was kept, the prediction performance of $K$ = 6 was better than that of no decision. This shows that the introduction of agent behavior decision is more helpful for prediction in a single modality, whereas the decision information only contains one positive label. Lane decision is helpful for multi-modal trajectory prediction as there may be multiple lane for attention. The proposed model employs both decisions, which act as a regulator, improving model predictive performance in both uni- and multi-modals simultaneously.

## 5 CONCLUSIONS

Trajectory prediction and behavioral decision-making are often considered as two separate tasks, although they are closely related. In addition, simple cue connection poses difficulties in exploring the internal relationships between multiple cues. In this study, we proposed the HT joint neural network, which combines multiple cues for trajectory prediction and decision-making. The network consists of an encoder, a feature selection network, and a decoder. The proposed network reduces the influence of perceptual noise using three attention mechanisms as well as organically selects and fuses features for decoding. On the Argoverse dataset, the comprehensive performance of the proposed model was better than those of other trajectory prediction models. The comparison and ablation experiment results show that the proposed model can better fuse multiple cues. Additionally, HT successfully outputs behavior and lane decisions simultaneously for intelligent vehicle path planning and control. We demonstrated the visualization results of a lane-scaled dot-product attention matrix, proving that lane decision attention adequately mines the multimodality of agents and improves model interpretability. We also found that if there is an obviously large error in the input data of one frame, its trajectory can be smoothed through preprocessing; otherwise, it can be directly eliminated to reduce its impact. Lane decisions contribute to mining the multimodality of agents, and agent behavior decisions are helpful in single modal predictions. In addition, the proposed network has good practical applicability. The proposed model can output the trajectories of multiple modals and their probabilities, thus providing a more accurate solution space for subsequent motion planning. At the same time, the proposed model can output behavioral decisions, reduce computational overheads, and alleviate the shortage of computing resources.

Although the method proposed in this study performs well, there are some limitations. Presently, HT predictions are limited to 3 s. Longer-term trajectories should also be predicted. There is also an imbalance in the driving maneuver classes; the system lacks adaptive weights to compensate for each unbalanced loss function. Future research should be conducted on longer trajectory prediction tasks, and the results should be compared to the current ones. The class imbalanced loss function and its adaptive loss form should also be more closely examined.

FUNDING

DECLARATION OF COMPETING INTEREST

CREDIT AUTHORSHIP CONTRIBUTION STATEMENT